\documentclass[runningheads]{llncs}
\usepackage{graphicx, lmodern, amsmath, bm, booktabs, multirow, xpatch, hyperref, cleveref, url, enumerate, enumitem, xcolor}
\usepackage{caption}
\usepackage{esvect}
\usepackage{subcaption}
\usepackage[T1]{fontenc}
\usepackage[numbers, sort&compress]{natbib} 
\usepackage{appendix}
\setcitestyle{numbers,square}
\usepackage[export]{adjustbox}
\usepackage{marvosym}

\crefname{figure}{fig.}{figures}
\Crefname{figure}{Fig.}{Figures}
\Crefname{equation}{Eq.}{Equation}

\hypersetup{
    colorlinks=true,           
    linkcolor=blue,            
    citecolor=blue,            
    urlcolor=blue,             
    bookmarks=true,
    bookmarksopen=true,
    pdftitle={Aligned Anchor_Groups_Guided_Line_Segment_Detector},
    pdfauthor={Zeyu Li}
    pdfpagemode=FullScreen,
    }

\xpatchcmd{\thebibliography}{\section}{\section}{}{}
\makeatletter
\renewcommand{\thebibliography}[1]{%
  \section*{References}%
  \list{\@biblabel{#1}}{}%
}
\makeatother

\begin{document}
\title{Aligned Anchor Groups Guided Line Segment Detector}
%
%
\author{Zeyu Li \and Annan Shu\inst{(}\textsuperscript{\raisebox{-0.18ex}{\scalebox{1.4}{\Letter}}}\inst{)}}
%
\authorrunning{Z. Li \and A. Shu}
%
\institute{Department of Intelligent Manufacturing, CATL, Ningde 352100, Fujian, China \\
\email{\{lizy51, shuan01\}@catl.com}}

\maketitle              
\begin{abstract}
This paper introduces a novel line segment detector, the \textbf{A}ligned \textbf{A}nchor \textbf{G}roups guided \textbf{L}ine \textbf{S}egment \textbf{D}etector (AAGLSD), designed to detect line segments from images with high precision and completeness. The algorithm employs a hierarchical approach to extract candidate pixels with different saliency levels, including \textit{regular anchors} and \textit{aligned anchor groups}. AAGLSD initiates from these aligned anchor groups, sequentially linking anchors and updating the currently predicted line segment simultaneously. The final predictions are derived through straightforward validation and merging of adjacent line segments, avoiding complex refinement strategies. AAGLSD is evaluated on various datasets and quantitative experiments demonstrate that the proposed method can effectively extract complete line segments from input images compared to other advanced line segment detectors. The implementation is available at \url{https://github.com/zyl0609/AAGLSD}.

\end{abstract}
\section{Introduction}
\label{sec:intro}

Line segment detection, a fundamental task in computer vision, has attracted considerable research interests over the years\cite{desolneux2000meaningful,von2008lsd,akinlar2011edlines,suarez2022elsed,cho2017novel,almazan2017mcmlsd,ZHANG2021107834,lin2023level,9008267,xue2020holistically,10243120,Pautrat_Lin_2021_CVPR,DAI20221}. As low-level features in digital images, line segments are essential for various high-level vision tasks and applications, such as object detection\cite{TANG2022108885}, 3D reconstruction\cite{9010693} and SLAM\cite{Hirose2012FastLD,gomez2019pl}.

Current line segment detectors can be primarily categorized into: (1) handcrafted methods\cite{von2008lsd,akinlar2011edlines,cho2017novel,almazan2017mcmlsd,ZHANG2021107834,suarez2022elsed,lin2023level}; (2) learning-based methods\cite{9008267,xue2020holistically,10243120,Pautrat_Lin_2021_CVPR,DAI20221}. Although learning-based line segment detectors have garnered significant attention and have achieved remarkable performance on datasets\cite{coughlan2003manhattan,cho2017novel,hpatches_2017_cvpr,wireframe_cvpr18}, they are constrained by the scale of the datasets and struggle to generalize effectively to common scenarios. Furthermore, the substantial computational cost required for inference presents challenges when deploying these models in environments with limited computing and storage capabilities. Therefore, handcrafted line segment detectors still remain highly significant in research.

In digital images, pixels within a line segment exhibit similar gradient magnitudes and orientations within their local neighborhoods. Despite disturbances such as noise and occlusions that can cause abrupt gradient changes, we observe that \textit{the salient parts of a line segment still maintain a high degree of gradient consistency} (i.e., aligned gradient orientations and high magnitudes). However, \textit{previous handcrafted line segment detectors have not explicitly made full use of this information}. \textbf{Motivated} by this, we introduce a hierarchical anchor extraction strategy to actively leverage pixels with different saliency levels as cues for line segment detection, classifying pixels into three categories: \textit{non-anchor}, \textit{regular anchor}, and \textit{aligned anchor group}. The proposed AAGLSD treats detection as a process of linking anchors, with the final predictions generated through a straightforward validation and merging process. The main contributions of this paper are as follows:
\begin{itemize}
    \item[$\bullet$] It presents a novel hierarchical anchor extraction strategy. Leveraging the gradient characteristics of rasterized line segments, it hierarchically classifies pixels into candidate points of varying saliency levels.
    \setlength{\itemsep}{1pt}
    \item[$\bullet$] It employs the pixel-routing process to select the next pixel and continuously links anchors to form line segments, and the detected line segments are further refined by applying straightforward validation and merging.
    \setlength{\itemsep}{1pt}
    \item[$\bullet$] The experiments demonstrate that the proposed method could effectively detect line segments with high precision and completeness compared to other advanced methods, achieving an excellent trade-off between precision and recall in structured scenes. Furthermore, it exhibits less sensitivity to changing illumination compared to two handcrafted baseline methods\cite{von2008lsd,akinlar2011edlines}.
\end{itemize}

\section{Related Work}
\label{related}
\subsection{Handcrafted Method}
Handcrafted line segment detectors mainly exploit local image features and the characteristics of line rasterization, grouping pixels within local neighborhoods according to predefined strategies and validating candidate line segments using statistical methods, thereby generating the final predicted line segments. 

LSD\cite{von2008lsd}, regarded as the baseline method, identifies line support regions and their minimum bounding rectangles by leveraging level-lines\cite{desolneux2000meaningful}. Validation of these regions is performed using an a-contrario model based on the Helmholtz principle, which effectively suppresses false positives in complex texture areas. The final predicted line segments are generated through a refinement process to these rectangles, ensuring accuracy and robustness.

EDLines\cite{akinlar2011edlines} operates on the edge map produced by Edge Drawing\cite{TOPAL2012862,akinlar2012edpf}, enhancing the detection efficiency. It traverses $edgels$ and greedily adds them, generating candidate line segments via the least square method. The method employs the Helmholtz principle to suppress the number of false alarms. Based on this, ELSED\cite{suarez2022elsed} further refines the pixel-routing (edge drawing) process, improving efficiency and robustness.

Linelet\cite{cho2017novel} introduces $linelet$ representation to model intrinsic properties of rasterized line segments. It connects neighboring $linelets$ that satisfy predefined conditions and further refines the angles of the line segments using the undirected graphical model and pairwise potentials. The candidate line segments are validated using the Bayes theorem.

MCMLSD\cite{almazan2017mcmlsd} first detects globally optimal lines in the Hough domain. Subsequently, it analyzes detected lines in the image domain to localize the line segments responsible for the peaks in the Hough map. This process is formulated as a Markov chain, and the line segments are detected via dynamic programming in linear-time.

AG3Line\cite{ZHANG2021107834} incorporates a jump strategy that actively connects line segments separated by gradient discontinuities. This approach ensures the continuity of line segments and enhances detection performance. Additionally, AG3Line validates each line segment based on the distribution characteristics of the gradient magnitudes.

\subsection{Learning-based Method}
Learning-based line segment detectors are primarily trained on the Wireframe dataset\cite{wireframe_cvpr18} and are used for wireframe parsing. Moreover, due to outstanding performance in downstream visual tasks, self-supervised learning-based methods have also drawn increasing attention.

L-CNN\cite{9008267} proposes a two-stage end-to-end method for wireframe parsing. It first generates junction proposals from the feature map which is obtained through the backbone network, then a heuristic sample method is employed to generate positive and negative samples from the proposals and the ground-truth line segments. Finally, line segments are derived via the proposed LOI pooling layer and the verification module.

HAWP\cite{xue2020holistically} extends the L-CNN\cite{9008267} by introducing the holistic attraction field (HAT). It re-parameterizes sparse representation of a line segment into pixel-wise dense 4-D vectors. \cite{10243120} refines HAWP, presenting HAWPv2 for fully supervised training and HAWPv3 for self-supervised training. 

SOLD$\rm{^2}$\cite{Pautrat_Lin_2021_CVPR} proposes a deep network for joint line segment detection and description, introducing a self-supervised training method similar to \cite{8575521}. The resulting line segment descriptors are occlusion-aware and can consistently match line segments across different viewpoints.

F-Clip\cite{DAI20221} presents a one-stage line segment detector that represents each line segment by center, length, and angle, and directly predicts these parameters rather than generating a large number of line proposals and classifying them. In this way, it converts line segment detection into the pixel-wise classification and regression tasks.


\begin{figure}[tb]
    \centering
    \includegraphics[width=0.95\linewidth]{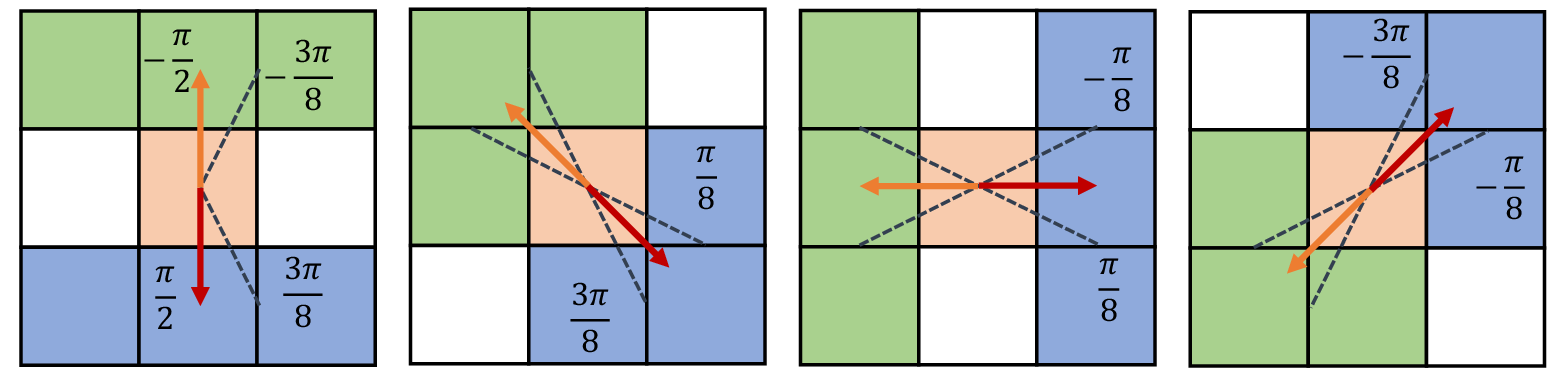}
    \caption{The search regions for pixel-routing, corresponding to various linking directions and the orientations of the fitting line. Blue grids indicate the candidate pixels associated with the $forward$ direction (red arrow), and green grids are associated with the $backward$ direction (orange arrow).}
    \label{fig:search neighborhood}
\end{figure}

\section{Approach}
\label{sec:algo}
In digital imagery, a line segment is rasterized and can be conceptualized as 8-connected components where pixels are sequentially linked in a specific orientation, characterized by aligned level-lines\cite{desolneux2000meaningful} and local magnitude maxima. The level-line and gradient magnitude are pivotal for line segment detection: the previous information provides the structural basis for a line segment, and the latter indicates the edges or boundaries. Both could facilitate to distinguish line segments from the background. Consequently, AAGLSD leverages both cues and introduces a hierarchical anchor extraction approach coupled with a skipping strategy. Through this hierarchical processing, different saliency levels are assigned to pixels. Line segments are extracted with high precision and completeness through the anchor-linking process, which links pixels of varying saliency to form a line segment and continuously refines the orientation of it. The AAGLSD algorithm is primarily composed of three components: (1) Hierarchical anchor extraction; (2) Linking anchors to form a line segment; (3) Line segment validation and merging.

\subsection{Hierarchical Anchor Extraction}
\label{ssec:anchor extraction}

Pixels constituting line segments typically exhibit particular relationships, such as aligned level-lines\cite{desolneux2000meaningful} and the presence of local gradient magnitude maxima. The detection of a continuous sequence of these pixels within a certain range strongly indicates their membership within a line segment. Therefore, this paper introduces a hierarchical anchor extraction approach that differentiates between low-level \textit{regular anchors}, which only exhibit local magnitude maxima, and high-level \textit{aligned anchor groups}, which also possess level-line consistency. This approach enhances the identification of meaningful and salient line segments.

In the context of line segment detection, an \textit{anchor} denotes a pixel with a high likelihood of belonging to a line segment and often appears in areas such as edges and complex textures within images. The process of extracting anchors begins with applying a 5$\times$5 Gaussian kernel to smooth the input grayscale image. Following this, the magnitude and orientation of the gradient are calculated according to \cite{von2008lsd}, denoted as $G_{mag}$ and $G_{ori}$, respectively. The level-line orientation, $\theta_l(P)$, for a pixel $P$, is derived as follows:
\begin{equation}
\theta_l(P)=G_{ori}(P)-\frac{\pi}{2},
\label{eq:level-line}
\end{equation}
where $\theta_{l}(P)$ spans the range $[-\frac{\pi}{2},\frac{\pi}{2})$, while $G_{ori}(P)$ ranges from $0$ to $\pi$. To reduce computational complexity and mitigate the impact of weak gradient responses, pixels with gradient magnitudes below the threshold $T_{mag}$ are eliminated. 

In edge drawing-based methods\cite{akinlar2011edlines,ZHANG2021107834,suarez2022elsed,lin2023level}, an anchor is identified when the gradient magnitude $G_{mag}(P)$ exceeds neighboring pixels within a search region by a threshold $T_{\text{anchor}}$. This paper introduces an expansion of search region types, as illustrated in \Cref{fig:search neighborhood}. It is observed that continuous pixels with local maximum gradient magnitude and aligned level-lines are more likely to constitute line segments. Based on this, a consistency check is employed to identify pixels that exhibit both maximum local gradient magnitudes and aligned level-lines with their neighbors. 

Given that pixels on salient line segments often reside in regions with significant gradient variation, pseudo-sorting following \cite{von2008lsd} is employed to prioritize the processing of pixels with high magnitudes. For a current pixel $P_i$, AAGLSD searches for the pixels $P_j$ and $P_k$ within its $n\times n$ neighborhood, both of which are local maxima within their respective regions. In the implementation, $n$ is set to 3, and the regions are depicted by the blue and green grids in \Cref{fig:search neighborhood}. The consistency check is then formulated as follows:
\begin{equation}
G_{mag}(P_i)-G_{mag}(P^{'}) \geq T_{anchor}, \label{eq:AAG-mag}
\end{equation}
\begin{equation}
    min\left( \pi-| \theta_{l}(P_i)-\theta_{l}(P^{'})|, |\theta_{l}(P_i)-\theta_{l}(P^{'})|\right) \leq T_{aligned}. \label{eq:AAG-ori}
\end{equation}
Here, $P^{'} \in \{P_j, P_k\}$, $T_{anchor}$ is the magnitude difference threshold, and $T_{aligned}$ is the tolerance difference between level-lines.

If \Cref{eq:AAG-mag} and \Cref{eq:AAG-ori} are met, the pixels $P_i$, $P_j$ and $P_k$ are collectively merged into a single \textit{aligned anchor group} (AAG). By iterating through all pixels, a set of high-level anchors, $\bf {AAGs}$, can be derived. It can be formulated as
$${\bf {AAGs}}=\{AAG_i=(P_i^0,P_i^1,P_i^2),i=0,1,...,M\}.$$
Subsequently, non-maximum suppression\cite{canny1986computational} is applied to the image, yielding a set of local maxima pixels:
$${\bf P}^{nms}=\{P_{i}^{nms}=(x_i,y_i), i=0,1,...,N\}.$$
By removing those pixels which are elements of ${\bf P}^{nms}$ but also found within $\bf{AAGs}$, the remaining elements are categorized as low-level \textit{regular anchors}. The pixels that are not assigned a saliency level are considered as \textit{non-anchor}.

\begin{figure}[tb]
\centering
    \rotatebox{90}{\#3}
    \begin{minipage}{.31\linewidth}
        \includegraphics[width=\textwidth]{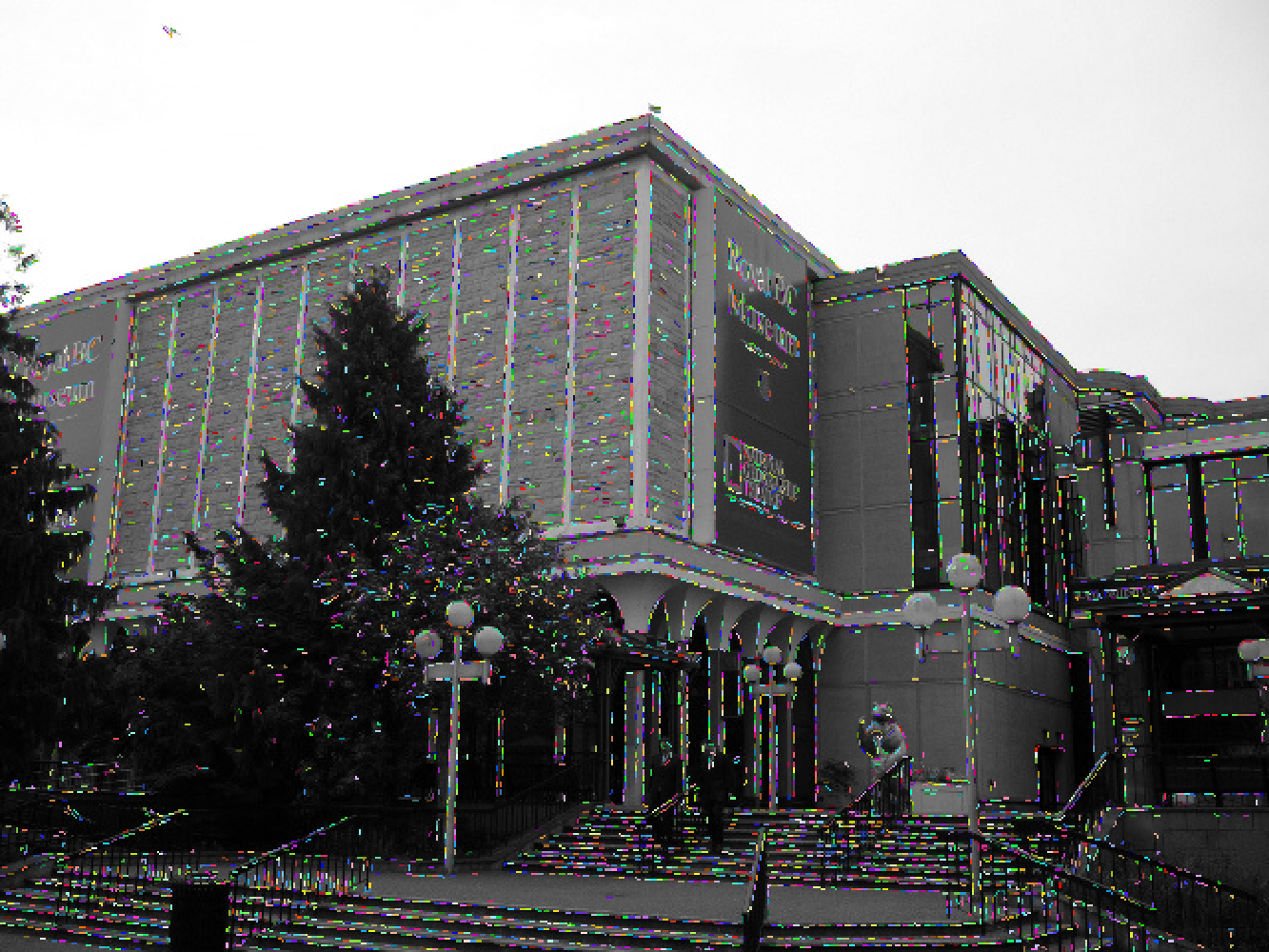}
    \end{minipage}%
    \hspace{1pt}
    \begin{minipage}{.31\linewidth}
        \includegraphics[width=\textwidth]{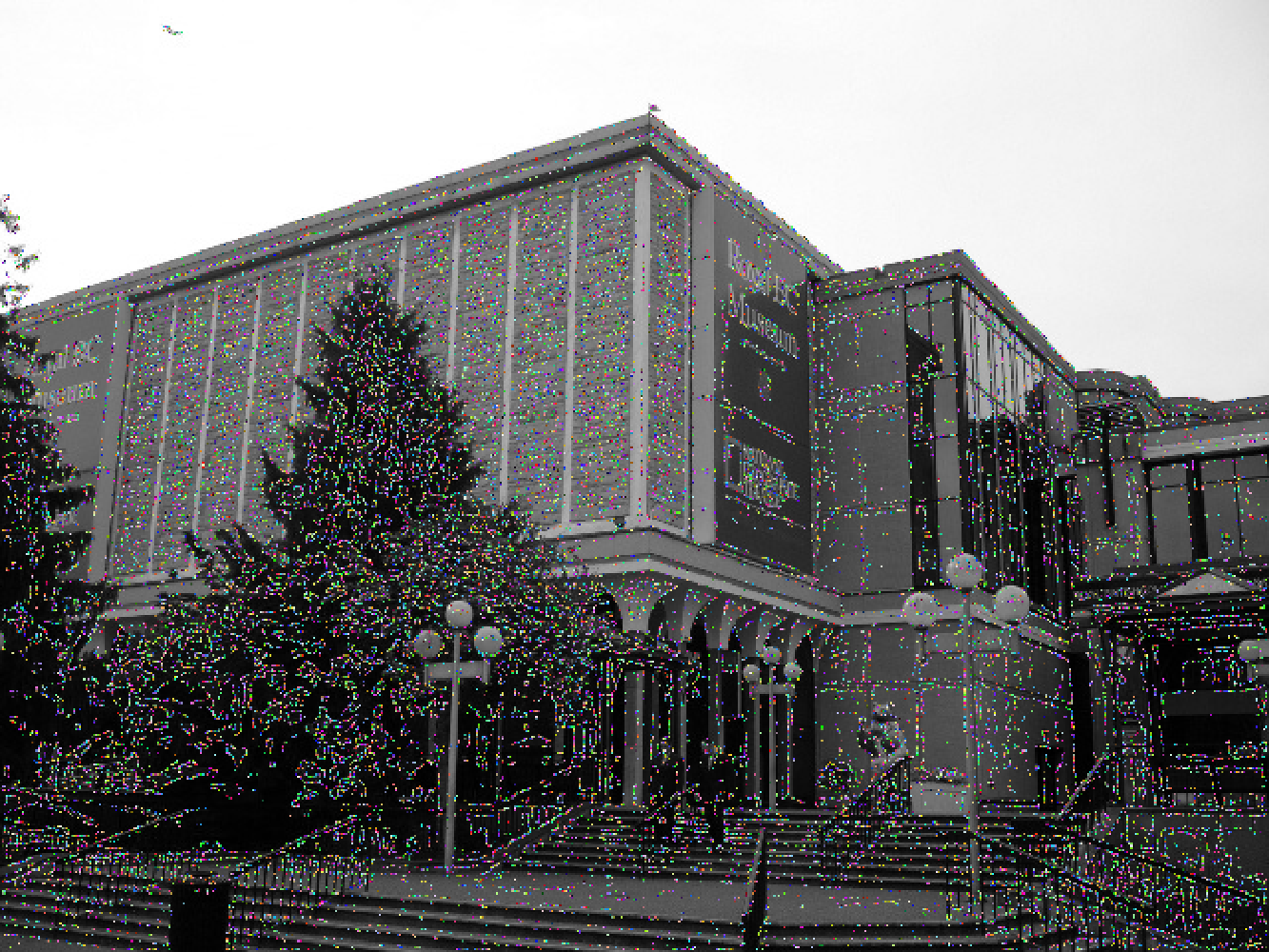}
    \end{minipage}%
    \hspace{1pt}
    \begin{minipage}{.31\linewidth}
        \includegraphics[width=\textwidth,frame]{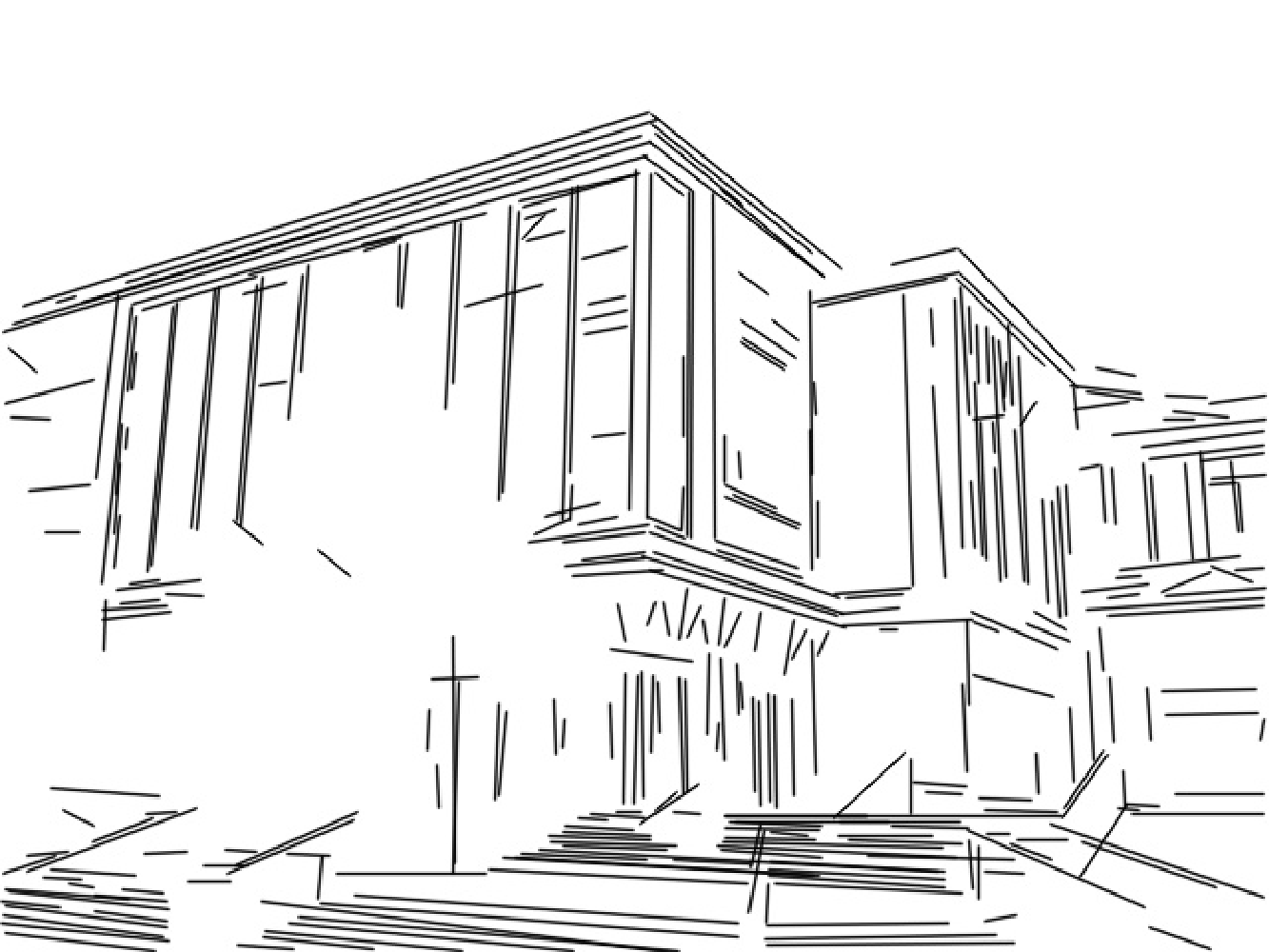}
    \end{minipage}%
    \vspace{1pt}
    \rotatebox{90}{\#11}
    \begin{minipage}{.31\linewidth}
        \includegraphics[width=\textwidth]{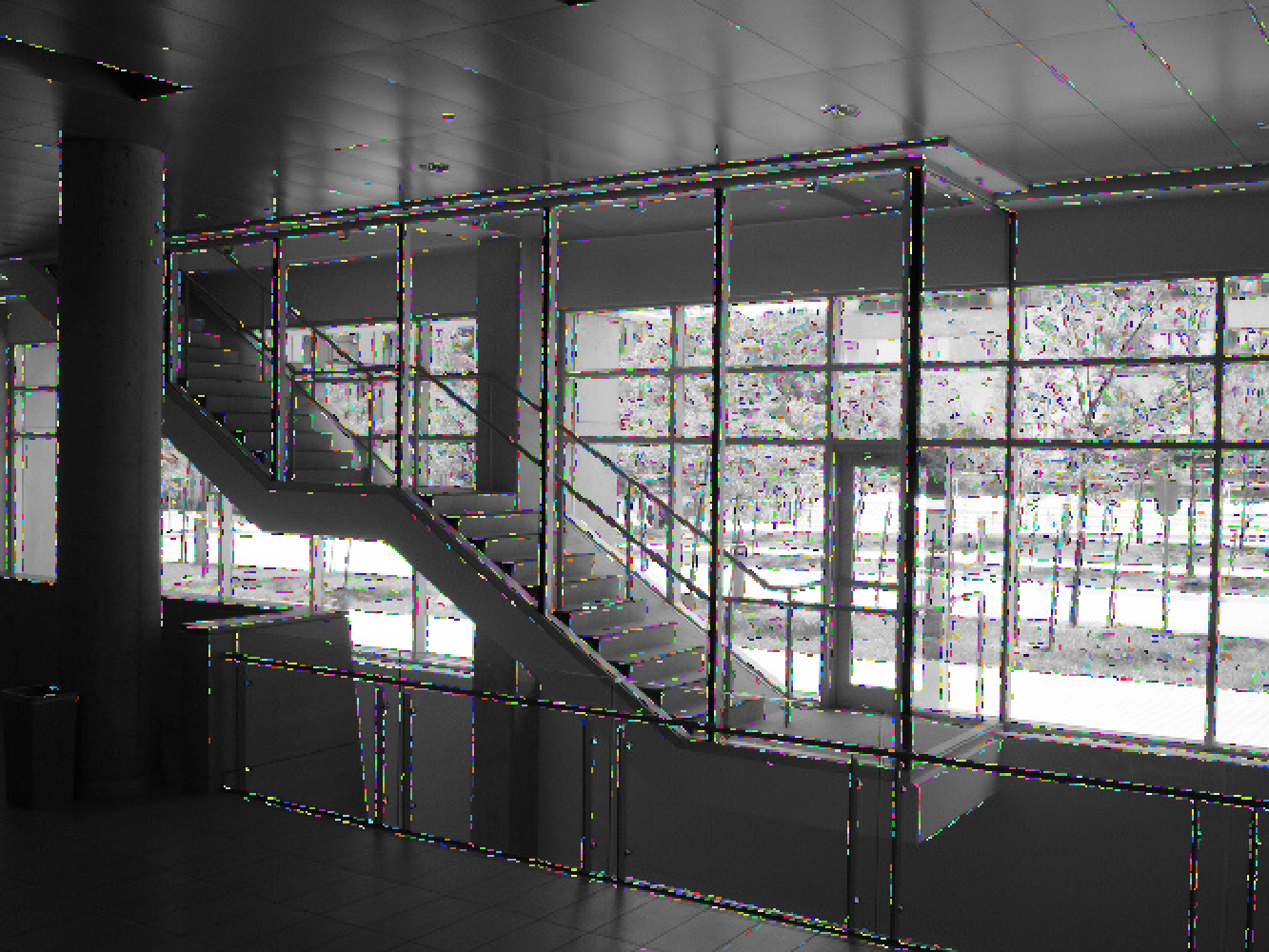}
    \end{minipage}%
    \hspace{1pt}
    \begin{minipage}{.31\linewidth}
        \includegraphics[width=\textwidth]{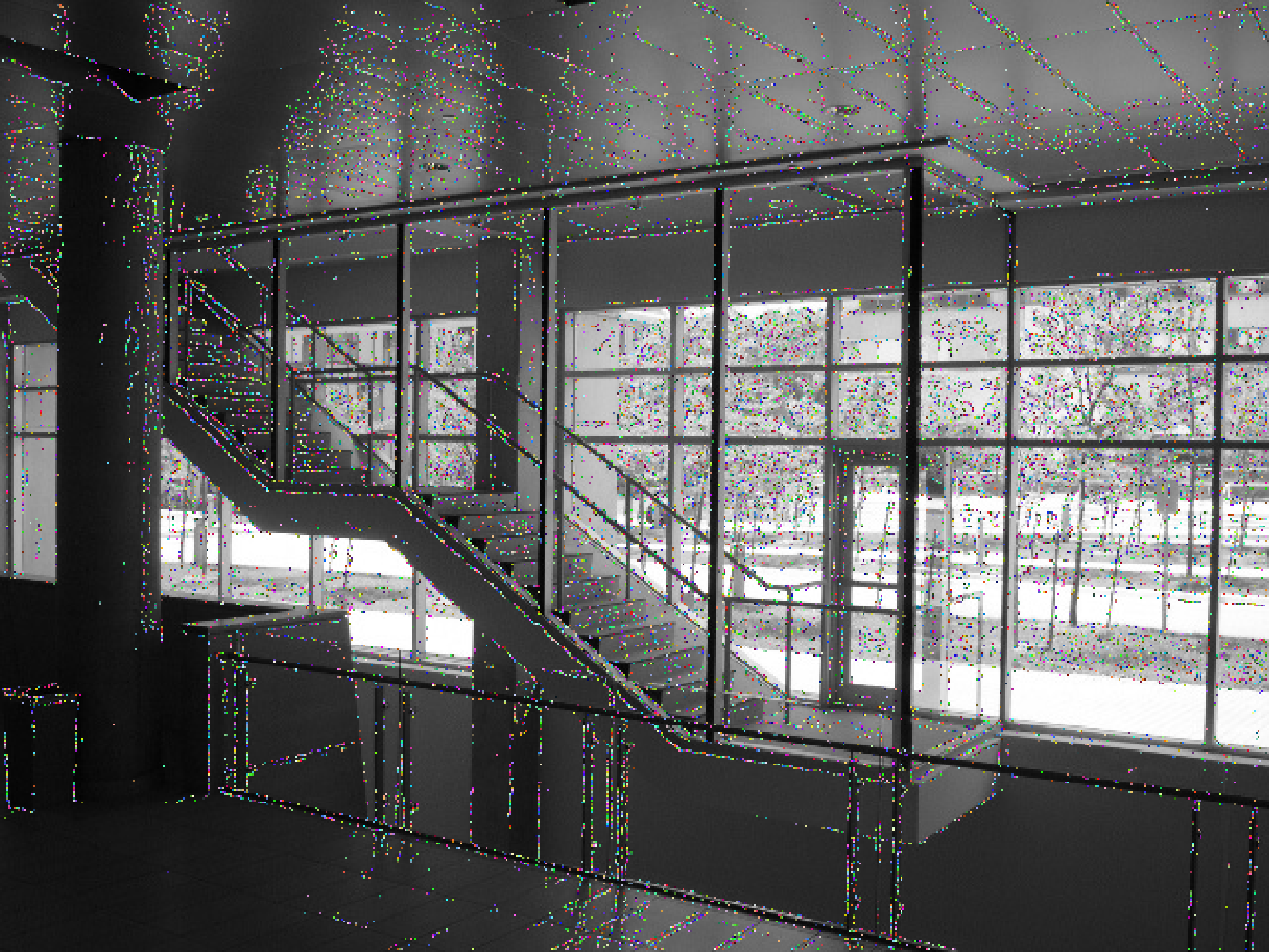}
    \end{minipage}%
    \hspace{1pt}
    \begin{minipage}{.31\linewidth}
        \includegraphics[width=\textwidth,frame]{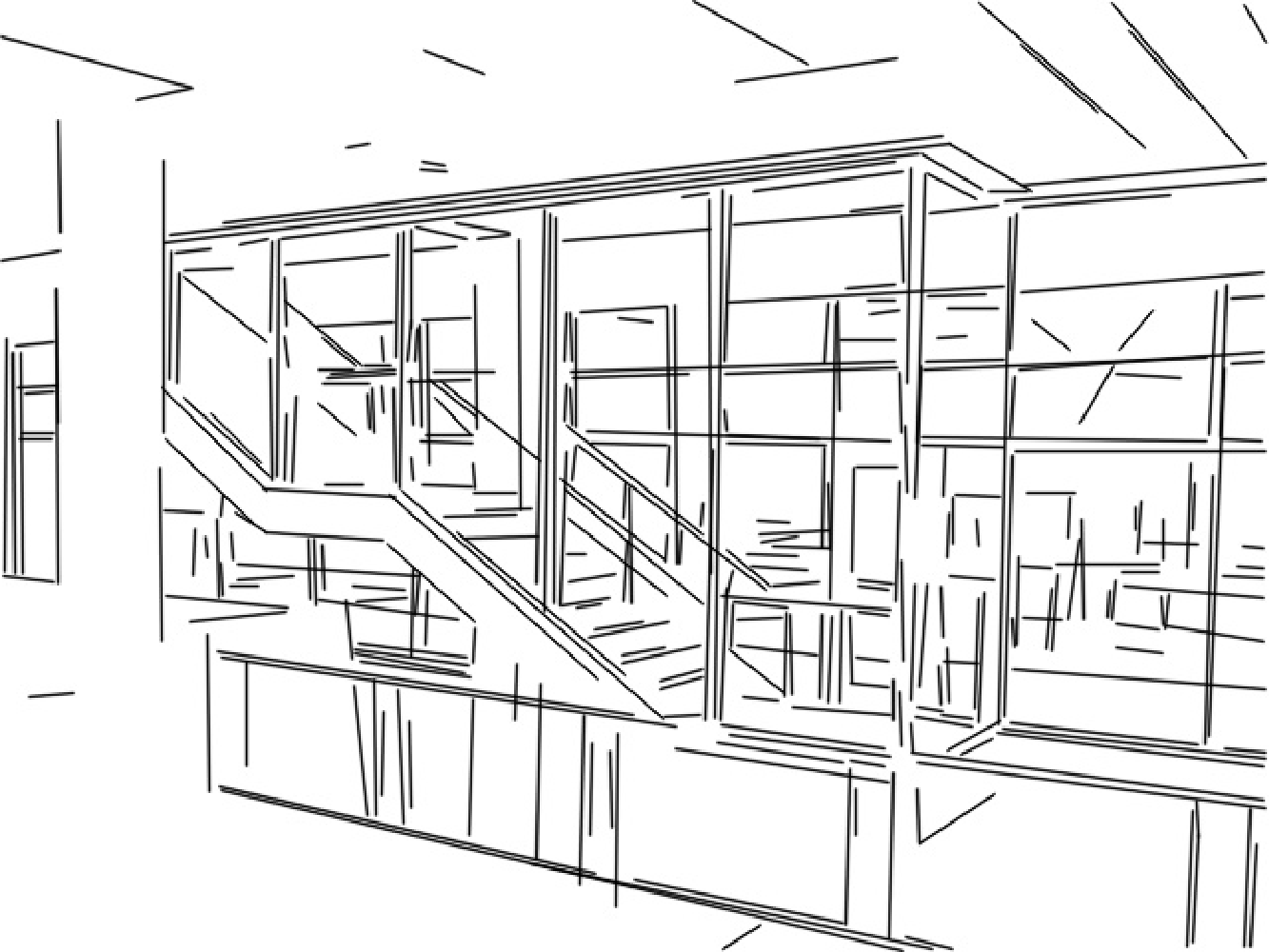}
    \end{minipage}%
    \vspace{1pt}
    \rotatebox{90}{\#94}
    \begin{minipage}{.31\linewidth}
        \includegraphics[width=\textwidth]{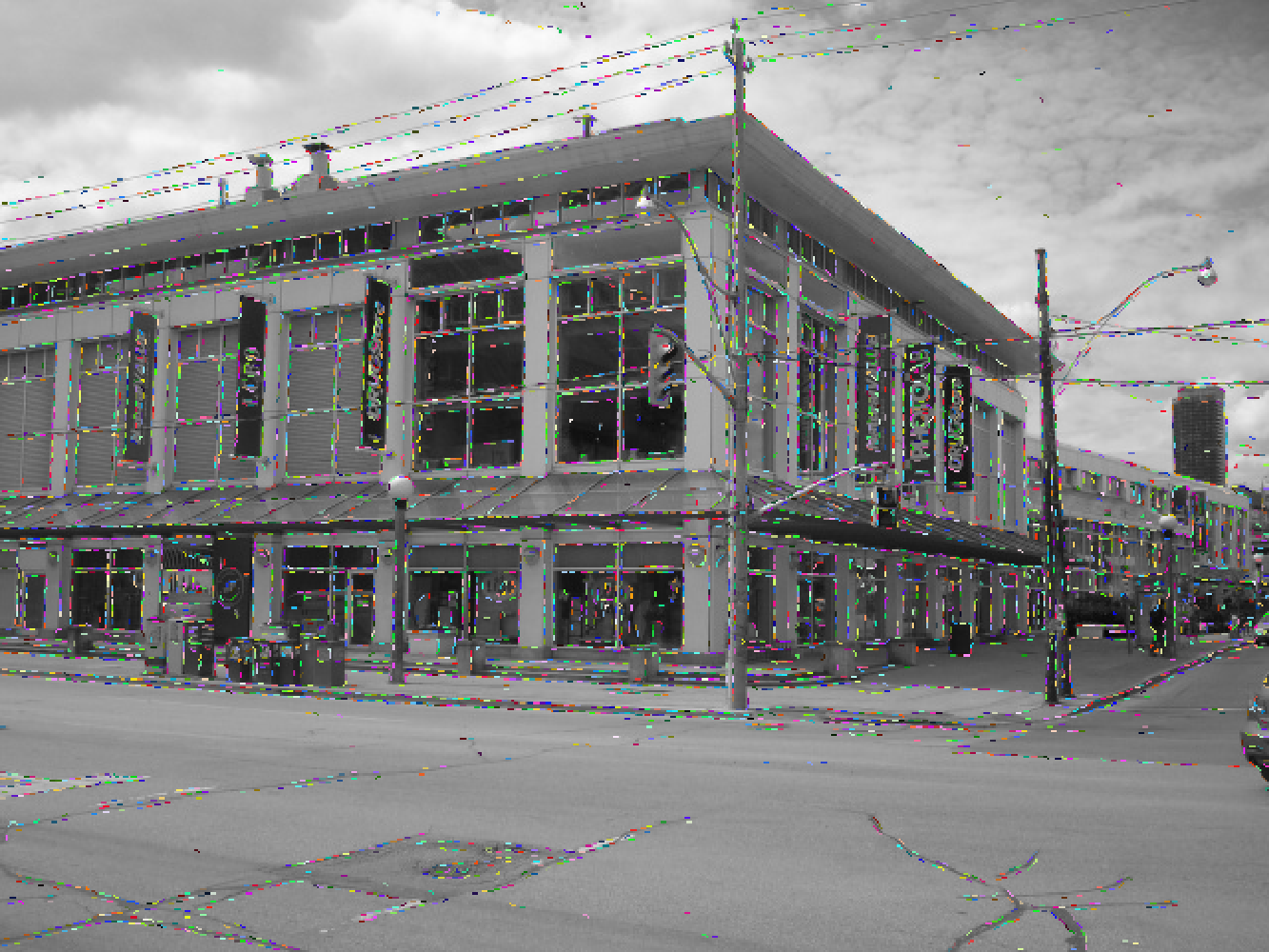}
        \vspace{-18pt}
        \caption*{(a)}
    \end{minipage}%
    \hspace{1pt}
    \begin{minipage}{.31\linewidth}
        \includegraphics[width=\textwidth]{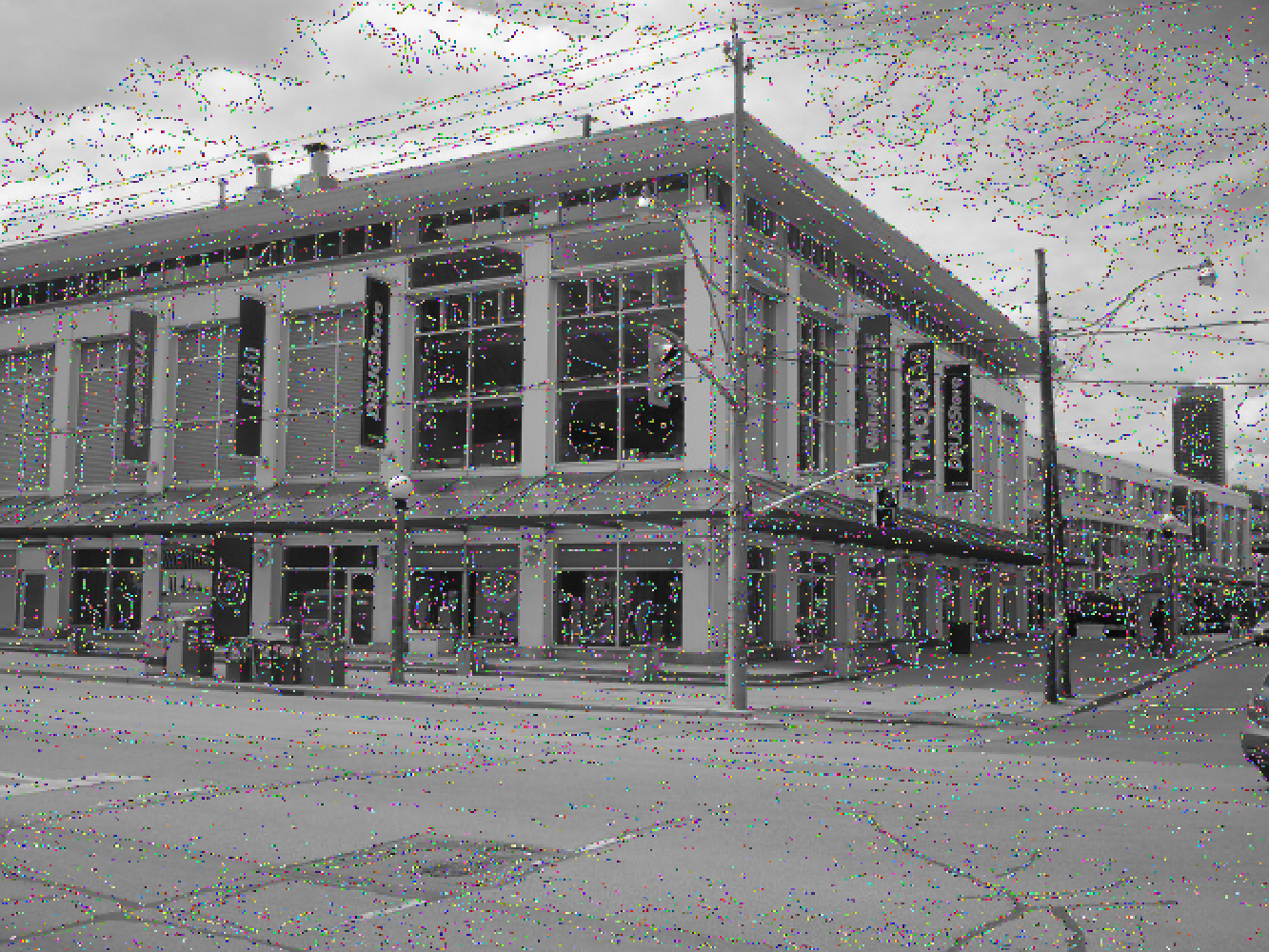}
        \vspace{-18pt}
        \caption*{(b)}
    \end{minipage}%
    \hspace{1pt}
    \begin{minipage}{.31\linewidth}
        \includegraphics[width=\textwidth,frame]{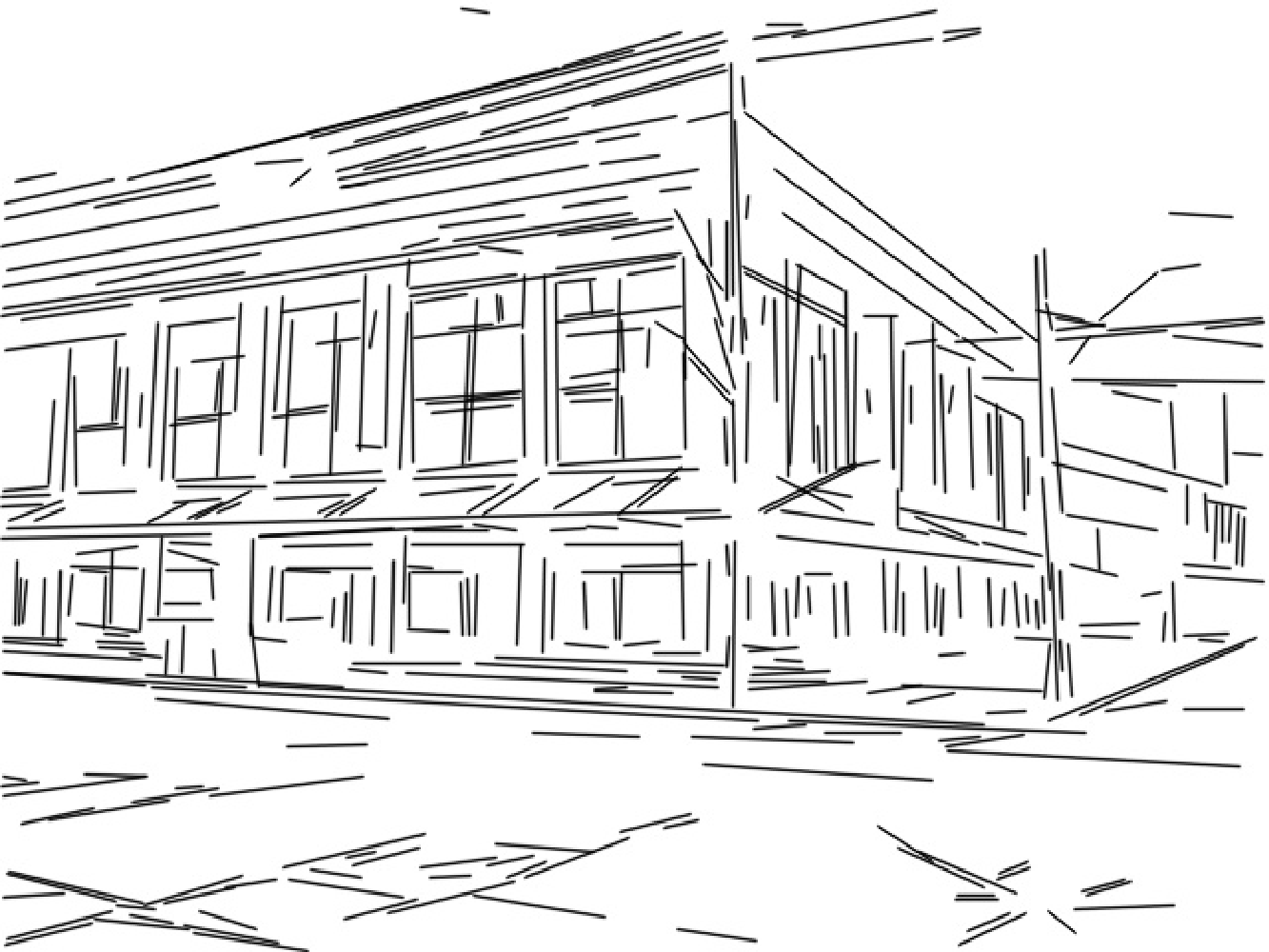}
        \vspace{-18pt}
        \caption*{(c)}
    \end{minipage}%
    \vspace{-5pt}
    \caption{Extracted hierarchical anchors and detected line segments by AAGLSD, where images are from the YorkUrban dataset\cite{coughlan2003manhattan}. From left to right: (a) aligned anchor groups; (b) regular anchors; (c) detected line segments.}
    \label{fig:result show}
\end{figure}

After this hierarchical processing, pixels are assigned varying levels of saliency, as shown in \Cref{fig:result show}. For an AAG, the consistency of gradient implies a high likelihood of being on a rasterized line segment. It also provides information about the general orientation of the line segment they may belong to, guiding subsequent line segment detection. For a low-level regular anchor, it is merely individual pixel that is local maxima, serving only as supplementary information. Non-anchor pixels are deemed \textit{background} elements and are ignored.

\begin{figure}[tb]
    \centering
    \includegraphics[width=0.8\linewidth]{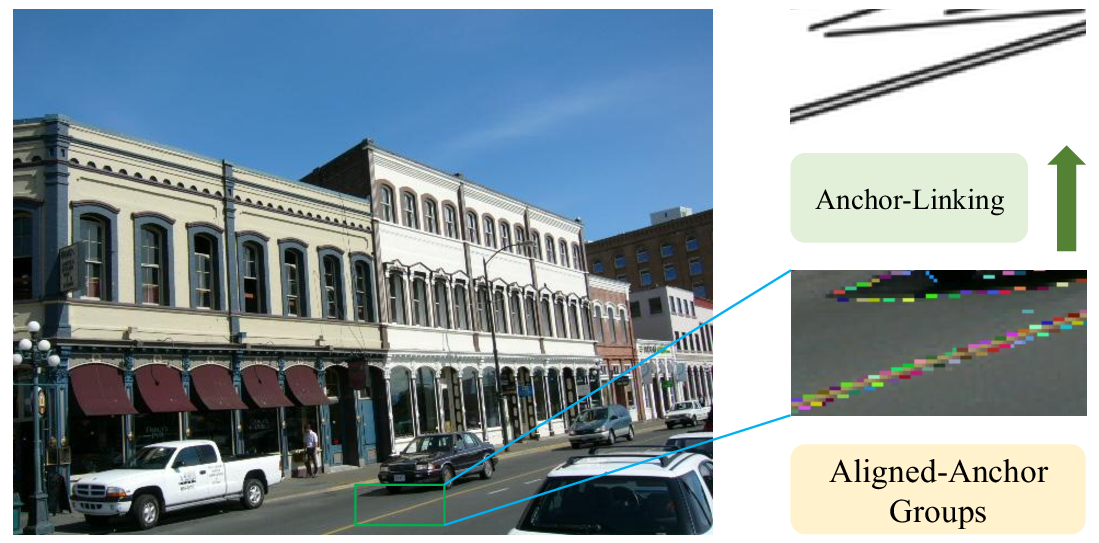}
    \caption{The linking process links AAGs progressively and form a line segment, where every three pixels with the same color is an AAG.}
    \label{fig:linking}
\end{figure}

\subsection{Linking Hierarchical Anchors}
\label{ssec:anchor linking}
In order to detect line segments in images using AAGs and regular anchors, AAGLSD treats line segment detection as an \textit{anchor-linking} process. This process iteratively starts from the AAG and employs pixel-routing to search for anchors that satisfy predefined criteria. These anchors are then greedily aggregated to form candidate line segments, with the search direction dynamically determined during the aggregation process.

The pixel-routing\cite{akinlar2011edlines,ZHANG2021107834,suarez2022elsed,lin2023level} selects the next pixel $P_{next}$, which features the maximum local gradient magnitude within a search region determined based on the linking direction and $\theta_{seg}$, as shown in \Cref{fig:search neighborhood}. To address discontinuities caused by noise and occlusions, a hierarchical skipping strategy, inspired by \cite{ZHANG2021107834,suarez2022elsed}, is introduced. Skipping steps are indicated as $S_{ra}$ for regular anchors and $S_{aag}$ for AAGs. The algorithm maintains a variable $remainSteps$ to determine the continuation or termination of linking, which depends on whether $remainSteps$ is greater than 0. Initially, $remainSteps$ is set to $S_{min}$.

The anchor-linking process begins with an \textit{available} AAG (i.e., the statuses $linked$ and $visited$ are false), and then extends in both $forward$ and $backward$ directions, progressively linking additional regular anchors and AAGs based on the orientation of the predicted line segment $\theta_{seg}$. \Cref{fig:linking} presents the line segments generated through this linking process. During the linking process, regular anchors or AAGs are incorporated into the set of pixels $\mathbf{P}_{fit}$ to refine $\theta_{seg}$. In addition, 
$P_{end}$ is used to determine the endpoint of the line segment in the $forward$ or $backward$ direction. The linking process is as follows:
\begin{itemize}
    \item[$\bullet$] If $P_{next}$ is a regular anchor, it is incorporated into $\mathbf{P}_{fit}$ while updating $remainSteps$ to $S_{ra}$.
    \setlength{\itemsep}{1pt}
    \item[$\bullet$] If $P_{next}$ is part of an available AAG and the AAG's level-line $\theta_{aag}$ is aligned with $\theta_l(P_{next})$, then all pixels of the AAG are added to $\mathbf{P}_{fit}$, $remainSteps$ is set to $S_{aag}$ and both statuses are updated to true. $\theta_{aag}$ is formulated as
    \begin{equation}
        \theta_{aag}=atan\left(\frac{\sum_{P \in AAG}{sin(\theta_l(P))}}{\sum_{P \in AAG}{cos(\theta_l(P))}}\right).
    \end{equation}
    \setlength{\itemsep}{1pt}
    \item[$\bullet$] If $P_{next}$ is within an available AAG but $\theta_{aag}$ is misaligned with $\theta_{seg}$, the linking process attempts to extend in the orientation of $\theta_{seg}$. Finding an available AAG within $S_{min}$ steps prompts moving $P_{next}$ to the center pixel of this AAG and performing the same steps as under the aligned condition.
    \setlength{\itemsep}{1pt}
    \item[$\bullet$] If $P_{next}$ is a non-anchor or part of an unavailable AAG, it suggests a potential discontinuity, and $remainSteps$ is decremented.
\end{itemize}
With each addition to $\mathbf{P}_{fit}$, the least squares method (LSM) is applied to refine $\theta_{seg}$, and $P_{end}$ is moved to $P_{next}$. Then, $P_{next}$ is determined by pixel-routing, and the linking process continues until $remainSteps$ is less than or equal to 0, or when the perpendicular distance from $P_{next}$ to the predicted line segment is greater than the distance tolerance $T_{dist}$.

In the initial state, $\mathbf{P}_{fit}$ consists of only three pixels of an AAG. Employing the LSM directly might lead to an unstable orientation. Therefore, AAGLSD initializes the orientation of the predicted line segment as
\begin{equation}
    \theta_{init}=\theta_{aag}.
\end{equation}
After linking $forward$ and $backward$, the terminal pixels, denoted as $P_{end}^f$ and $P_{end}^b$, are projected orthogonally onto the fitting line derived from $\mathbf{P}_{fit}$ using the LSM, representing the predicted line segment $L_{pred}$. Upon completion of traversing all AAGs and the linking process, the set of predicted line segments is determined.

\begin{figure}[htb]
    \centering
    \includegraphics[width=0.8\linewidth]{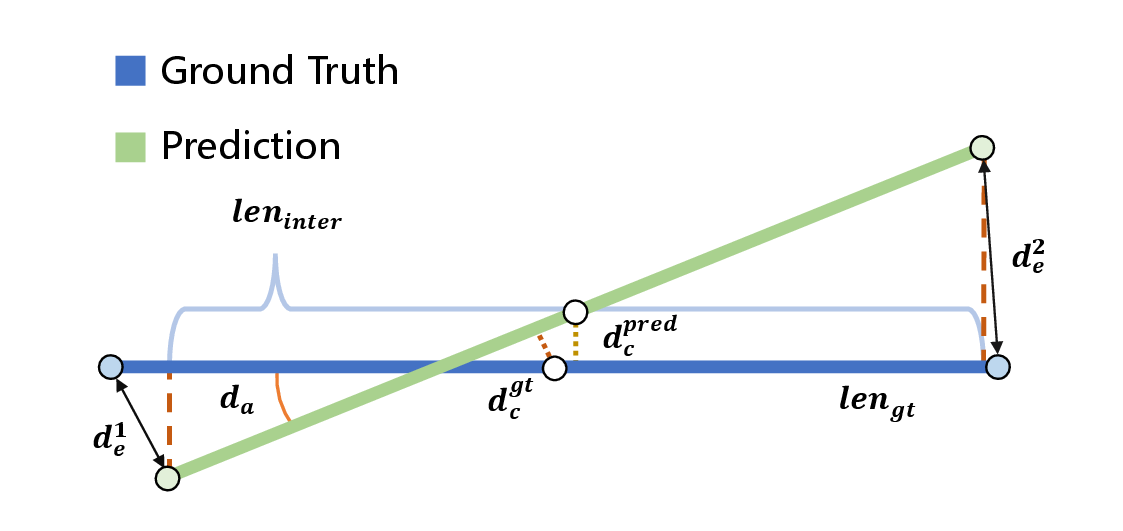}
    \caption{Parameters between the prediction and the ground truth. $d_a$ is the angle difference; $d_c^{pred}$ is the perpendicular distance from the midpoint of the predicted segment to the ground truth, while $d_c^{gt}$ is the opposite; $d_e^1$ and $d_e^2$ are the distances between the endpoints; $len_{gt}$ is the length of the ground truth; $len_{inter}$ is the length of the projection of predicted line segment onto the ground truth.}
    \label{fig:gt-pred}
\end{figure}

\subsection{Line Segments Validation and Merging}
\label{ssec:validate}
Handcrafted methods often encounter false positives in areas with high noise or sharp transitions. In contrast to previous works\cite{von2008lsd,akinlar2011edlines,almazan2017mcmlsd,cho2017novel,ZHANG2021107834} which validate line segments based on statistical models, e.g. the Helmholtz principle\cite{von2008lsd}, AAGLSD employs a straightforward validation approach. Moreover, AAGLSD merges the duplicate line segments to refine the predictions.

Line segments with a high density of anchors and alignments are more likely to be true positives. Thus, AAGLSD applies thresholds $T_{len}$, $N_{aag}$, $\rho_1$ and $\rho_2$, which correspond to the line segment's length, the number of AAGs, the density of anchors and the density of alignments, respectively. Only those line segments that exceed all these thresholds are retained. 

Due to factors such as lens defocus, edges may be blurred and occupy several pixels, which can lead to duplicate predictions of the same line segment. Since these duplicate line segments originate from the same structural information, merging these detected line segments is a more effective approach compared to directly removing them. To address this issue, the criteria for merging two line segments are formulated:
\begin{equation}
\left\{
    \begin{array}{l}
        dist_{c}^{max}(L,L_{pred}) \leq D_c\\
        dist_{a}(L,L_{pred}) \leq D_a\\
        L \cap L_{pred} \neq \phi,
    \end{array}
\right.\label{eq:cond1}
\end{equation}
or
\begin{equation}
\left\{
    \begin{array}{l}
        dist_{e}^{min}(L,L_{pred}) \leq D_e\\
        L \cap L_{pred} = \phi,
    \end{array}
\right.\label{eq:cond2}
\end{equation}
where $L$ is a candidate line segment and may be merged. $dist_{c}^{max}(\cdot,\cdot)$ and $dist_{a}(\cdot,\cdot)$ denote the maximum perpendicular center distance and the angular difference, and $dist_{e}^{min}(\cdot,\cdot)$ is the minimum distance between the endpoints, as illustrated in \Cref{fig:gt-pred}. $D_c$, $D_a$ and $D_e$ are the respective thresholds.

If \Cref{eq:cond1} or \Cref{eq:cond2} is met, both line segments are projected onto the image using the Bresenham algorithm, and the LSM is then applied to refine $\theta_{seg}$. The two pixels generated by the Bresenham algorithm with the maximum perpendicular distances to the new fitting line are then projected, representing the endpoints of the merged line segment. \Cref{fig:result show} presents the final line segments detected in some images from the dataset\cite{coughlan2003manhattan}.

\section{Experiment}
\label{sec:experiment}
This section presents quantitative experiments to demonstrate the effectiveness of AAGLSD. The performance of a line segment detector is evaluated through pixel-level metrics. To evaluate the quality of the detected line segments, a comparison of AAGLSD with other advanced line segment detectors is conducted on annotated datasets. Furthermore, the robustness of AAGLSD is evaluated with the two most popular handcrafted methods.

\subsection{Evaluation Datasets and Metrics}
\label{ssec:benchmark}

\subsubsection{Evaluation Datasets}
We utilize the YorkUrban dataset (YUD)\cite{coughlan2003manhattan}, which includes 102 images of indoor and outdoor scenes with the corresponding manual annotations for both vanishing point detection and line segment detection. Furthermore, the extended version of this dataset, the YorkUrban-LineSegment dataset (YULD)\cite{cho2017novel}, which features refined endpoints and complete annotations of line segments, is incorporated. We also conduct experiments on the HPatches dataset\cite{hpatches_2017_cvpr} which includes images captured under various illumination conditions and from different viewpoints. 

\subsubsection{Evaluation Metrics}
A line segment detector is evaluated in two aspects. First, we use average precision (AP), average recall (AR), intersection of union (IoU) and F-Score to evaluate its capability in structured scenes (e.g., rooms), following the evaluation framework\cite{cho2017novel}. They could be formulated as
\begin{equation}
    AP=\frac{\sum_i\sum_j|L_{tp}^i \cap L_{gt}^j|}{\sum_k|L_{pred}^k|},
\end{equation}
\begin{equation}
    AR=\frac{\sum_i\sum_j|L_{tp}^i \cap L_{gt}^j|}{\sum_k|L_{gt}^k|},
\end{equation}
\begin{equation}
    IoU=\frac{\sum_i\sum_j|L_{tp}^i \cap L_{gt}^j|}{\sum_i\sum_j|L_{tp}^i \cup  L_{gt}^j|},
\end{equation}
\begin{equation}
    F-Score=\frac{2\cdot AP \cdot AR}{AP+AR}.
\end{equation}
Here, $L_{gt}$ represents the ground-truth line segment. To measure completeness, these metrics are applied with varying thresholds for the intersection area ratio $\lambda_{area}=len_{inter}/len_{gt}$, as depicted in \Cref{fig:gt-pred}. To ensure that over-connected line segments are not treated as true positives, following \cite{ZHANG2021107834}, $L_{pred}$ is considered a true positive $L_{tp}$ only if it satisfies both \Cref{eq:cond1} and the following condition:
\begin{equation}
        \frac{|L_{pred} \cap L_{gt}|}{|L_{gt}|} \geq \lambda_{area} \enspace and \enspace \frac{|L_{pred} \cap L_{gt}|}{|L_{pred}|} \geq \lambda_{area}.
        \label{eq:TP}
\end{equation}
Second, we measure its robustness by \textit{repeatability}\cite{awrangjeb2008robust,lin2023level}, which is formulated as
\begin{equation}
\label{eq:rep}
    repeatability=\frac{N_{match}}{2} \cdot (\frac{1}{N_{ref}} + \frac{1}{N_{test}}),
\end{equation}
where $N_{ref}$ and $N_{test}$ are the numbers of line segments detected in the reference image and the test image, respectively. $N_{match}$ represents the number of line segments in the test image that meet \Cref{eq:cond1} and \Cref{eq:TP} compared to the line segments detected in the reference image.

\begin{figure}[tb]
\centering
    \begin{subfigure}{\textwidth}
        \begin{minipage}{.33\linewidth}
            \includegraphics[width=\textwidth]{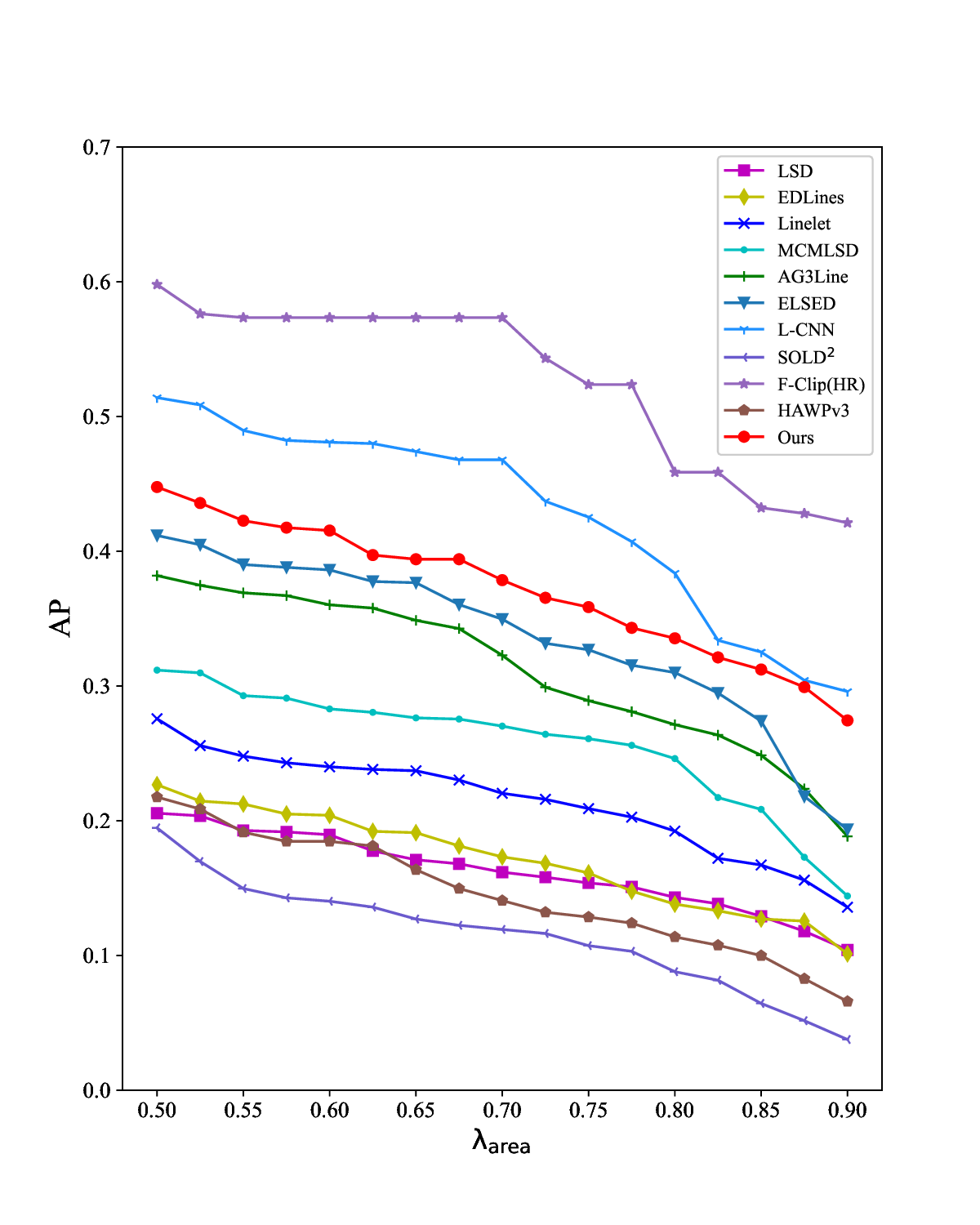}
            \label{fig:AP_YUD}
        \end{minipage}%
        \begin{minipage}{.33\linewidth}
            \includegraphics[width=\textwidth]{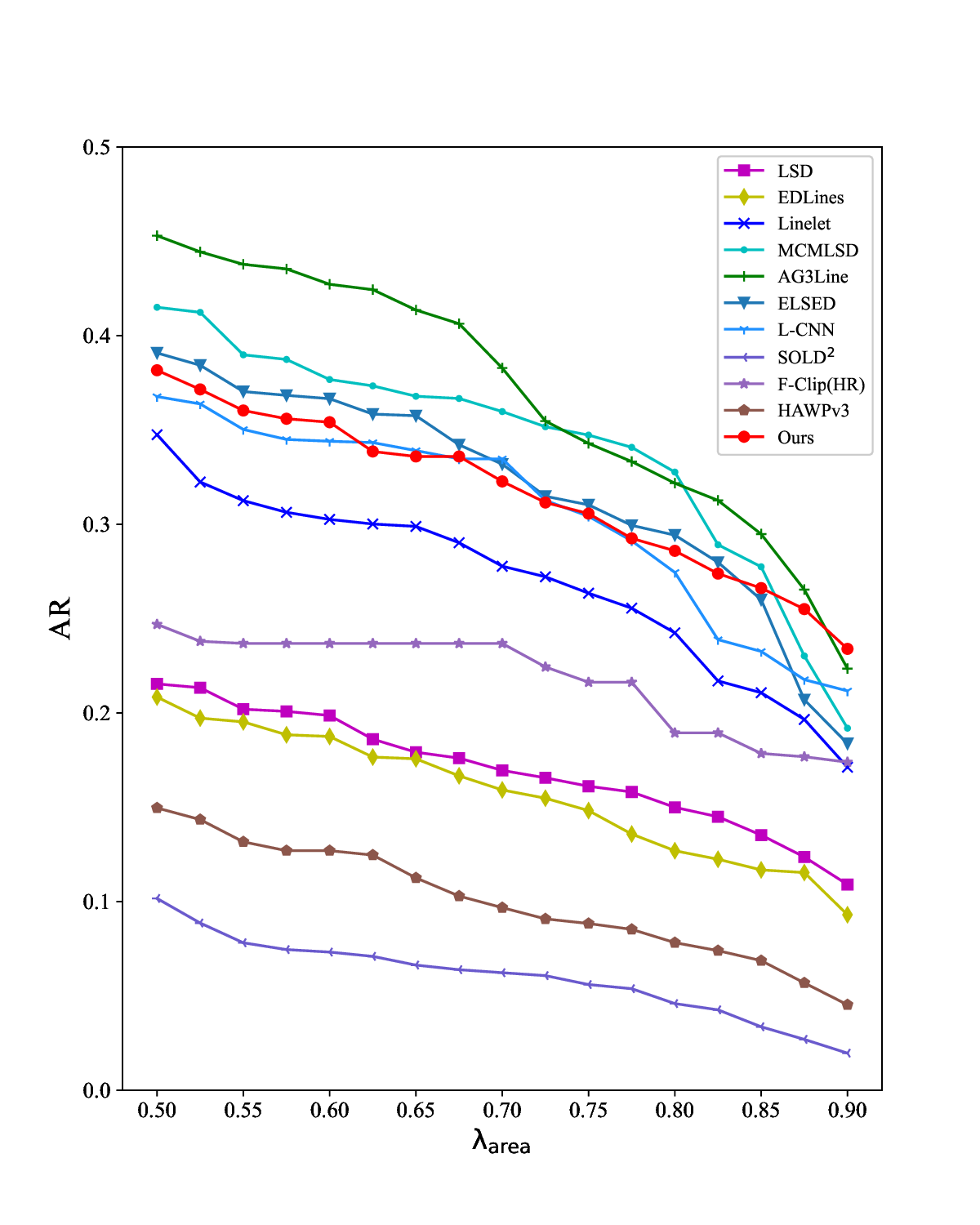}
            \label{fig:AR_YUD}
        \end{minipage}%
        \begin{minipage}{.33\linewidth}
            \includegraphics[width=\textwidth]{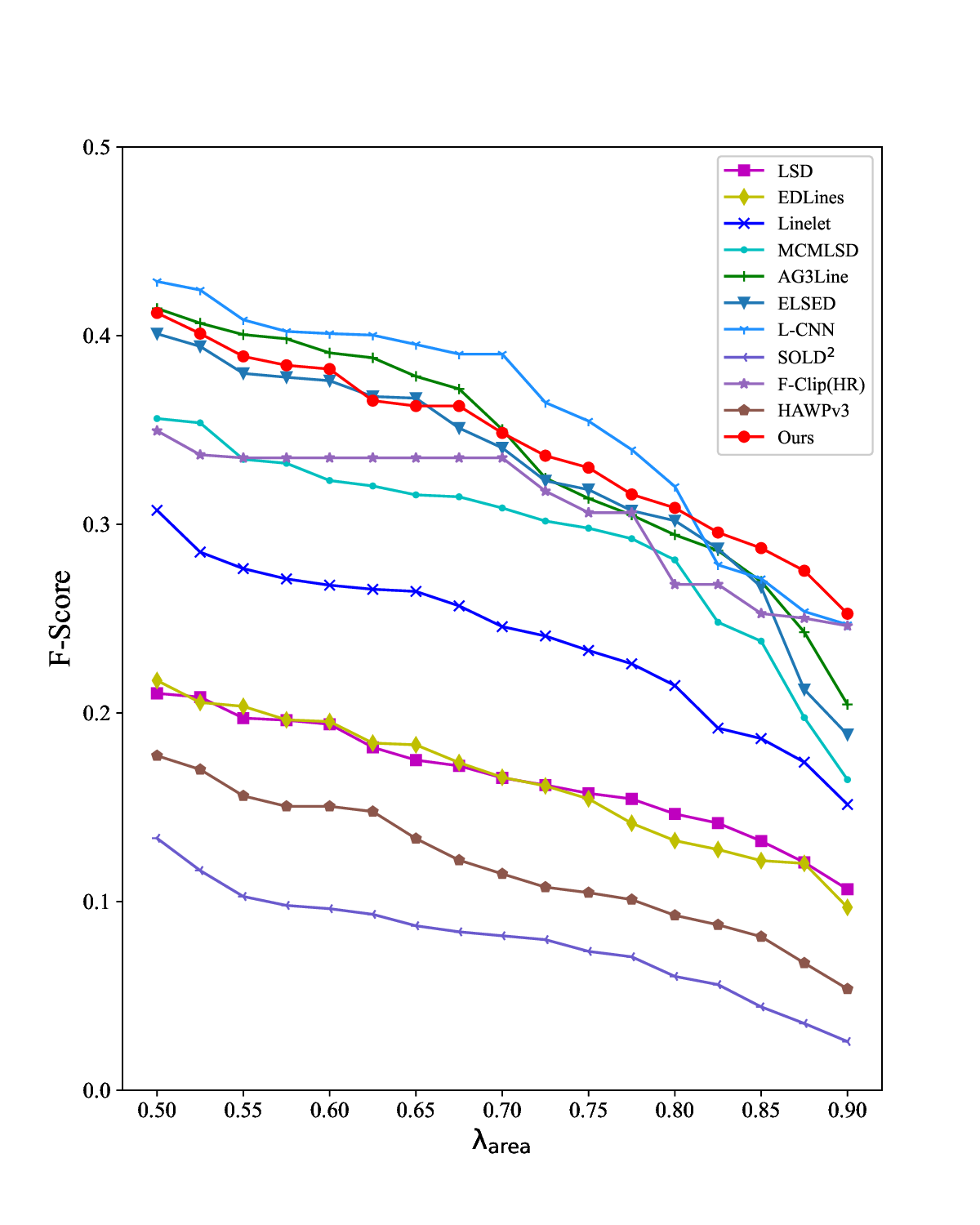}
            \label{fig:F_YUD}
        \end{minipage}\\%
        \vspace{-21pt}
        \caption{Metrics on the YorkUrban dataset.}
    \end{subfigure}
    \vspace{-3pt}
    \begin{subfigure}{\textwidth}
        \begin{minipage}{.33\linewidth}
            \includegraphics[width=\textwidth]{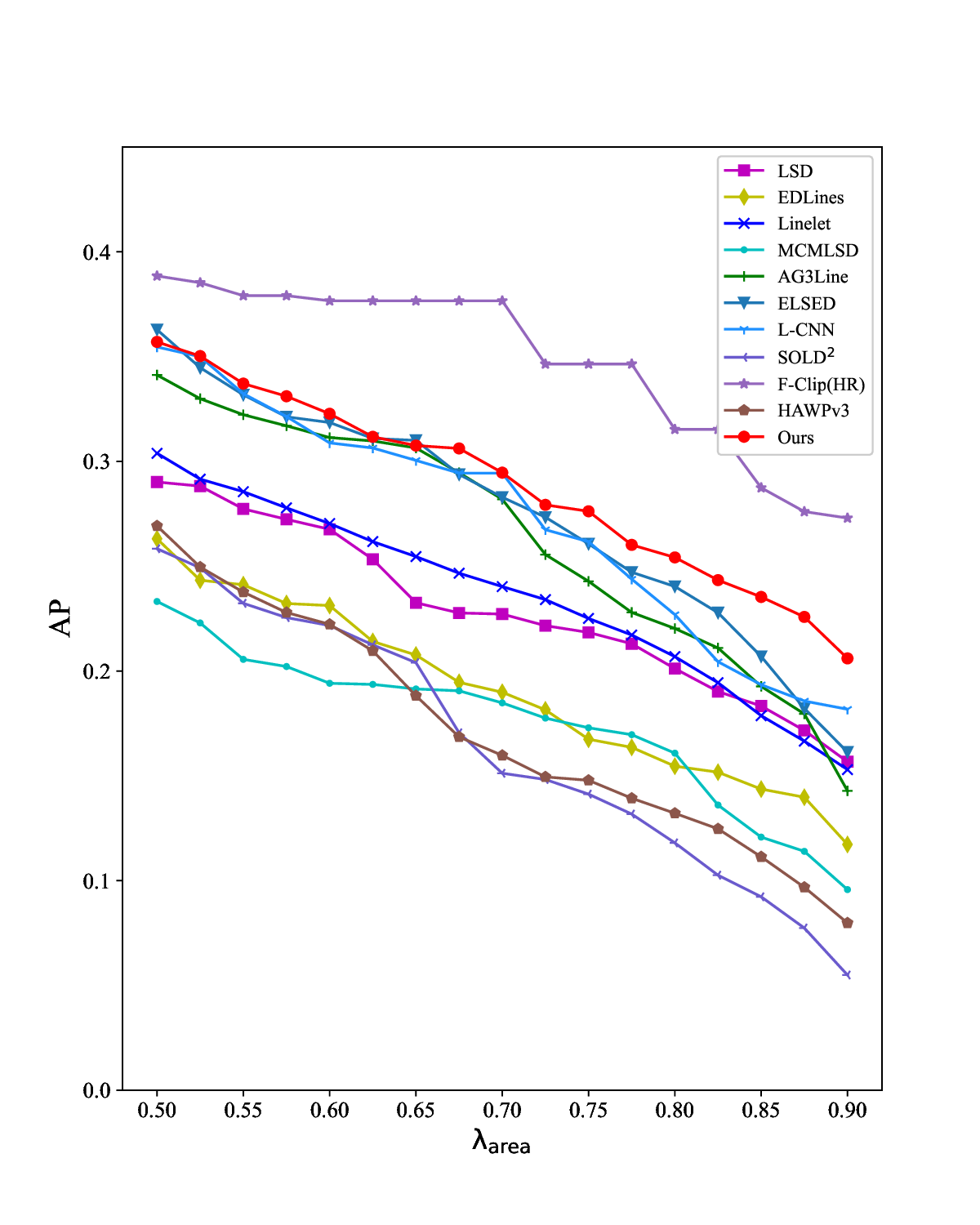}
            \label{fig:AP_YULD}
        \end{minipage}%
        \begin{minipage}{.33\linewidth}
            \includegraphics[width=\textwidth]{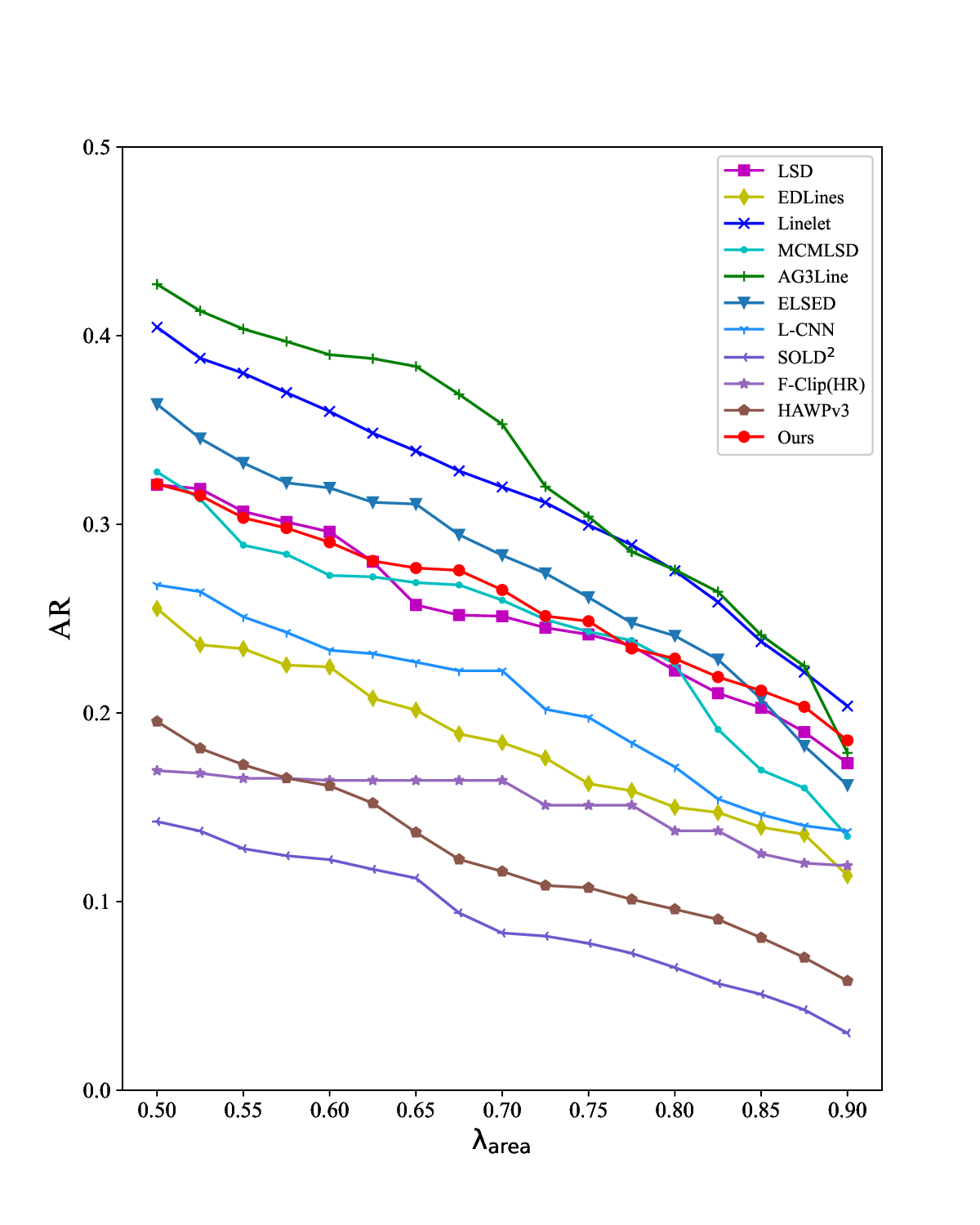}
            \label{fig:AR_YULD}
        \end{minipage}%
        \begin{minipage}{.33\linewidth}
            \includegraphics[width=\textwidth]{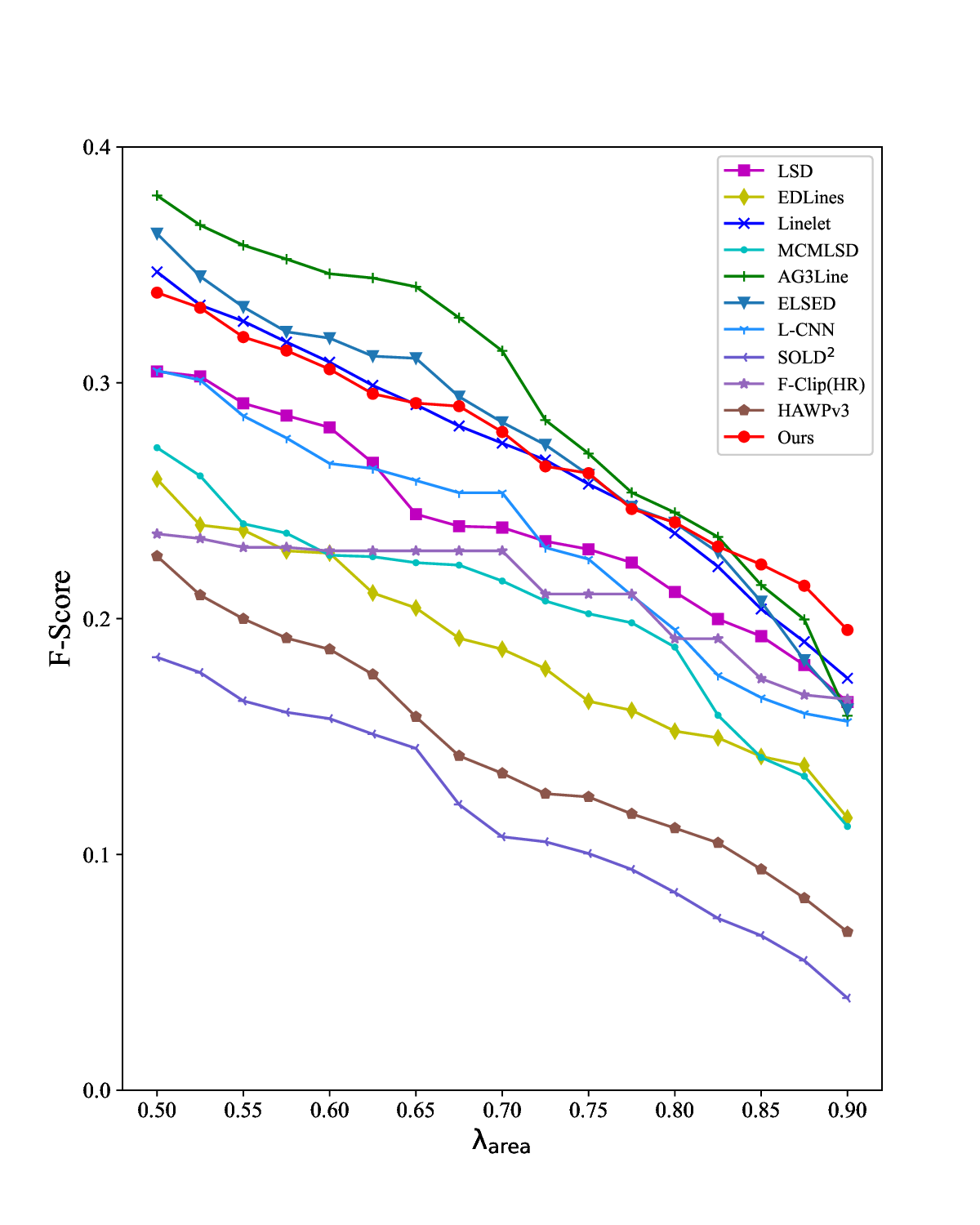}
            \label{fig:F_YULD}
        \end{minipage}\\%
        \vspace{-21pt}
        \caption{Metrics on the YorkUrban-LineSegment dataset.}
    \end{subfigure}
    \caption{Performance curves for line segment detectors evaluated on the YorkUrban dataset\cite{coughlan2003manhattan} and the YorkUrban-LineSegment dataset\cite{cho2017novel} in various $\lambda_{area}$.}
    \label{fig:curves}
\end{figure}

\begin{table}[tb]
    \centering
    \caption{Quantitative evaluation metrics in two conditions. Handcrafted methods are above the double line, while learning-based methods are below.}
    \begin{subtable}[t]{0.49\textwidth}
    \centering
    \caption{Metrics on the YUD\cite{coughlan2003manhattan}.}
    \label{tabel:1}
    \resizebox{\textwidth}{!}{
    \begin{tabular}{ccccccccc}
        \toprule
        \multirow{2}{*}{Method} & \multicolumn{2}{c}{$\rm{AP\uparrow}$} & \multicolumn{2}{c}{$\rm{AR\uparrow}$} & \multicolumn{2}{c}{$\rm{IoU\uparrow}$} & \multicolumn{2}{c}{$\rm{F-Score\uparrow}$} \\ 
        \cline{2-9} 
        & $\rm Con_1$ & $\rm Con_2$ & $\rm Con_1$ & $\rm Con_2$ & $\rm Con_1$ & $\rm Con_2$ & $\rm Con_1$ & $\rm Con_2$\\ 
        \midrule 
        LSD\cite{von2008lsd}  & 0.21 & 0.10  & 0.22 & 0.11 & 0.12 & 0.06 & 0.21 & 0.11  \\ 
        EDLines\cite{akinlar2011edlines}  & 0.23 & 0.10  & 0.21 & 0.09 & 0.12 & 0.05 & 0.22 & 0.10 \\ 
        Linelet\cite{cho2017novel}  & 0.28 & 0.14  & 0.35 & 0.17 & 0.18 & 0.08 & 0.31 & 0.15  \\ 
        MCMLSD\cite{almazan2017mcmlsd}  & 0.31 & 0.14  & 0.27 & 0.09 & 0.22 & 0.09 & 0.36 & 0.16  \\
        AG3Line\cite{ZHANG2021107834}  & 0.38 & 0.19  & \textcolor{red}{\textbf{0.45}} & \textcolor{green}{\textbf{0.22}} & \textcolor{green}{\textbf{0.26}} & \textcolor{blue}{\textbf{0.11}} & \textcolor{green}{\textbf{0.41}} & 0.20 \\
        ELSED\cite{suarez2022elsed}  & 0.41 & 0.19 & \textcolor{green}{\textbf{0.39}} & 0.18 & 0.25 & 0.10 &  0.40 & 0.19  \\
        \textbf{AAGLSD}(Ours)  & \textcolor{blue}{\textbf{0.45}} & \textcolor{blue}{\textbf{0.27}} & \textcolor{blue}{\textbf{0.38}} & \textcolor{red}{\textbf{0.23}} & \textcolor{blue}{\textbf{0.26}} & \textcolor{red}{\textbf{0.15}} & \textcolor{blue}{\textbf{0.41}} & \textcolor{red}{\textbf{0.25}} \\
        \hline
        \hline
        L-CNN\cite{9008267} & \textcolor{green}{\textbf{0.51}} & \textcolor{green}{\textbf{0.30}}  & 0.37 & \textcolor{blue}{\textbf{0.21}} & \textcolor{red}{\textbf{0.27}} & 0.14 & \textcolor{red}{\textbf{0.43}} & \textcolor{green}{\textbf{0.25}} \\
        SOLD$\rm{^2}$\cite{Pautrat_Lin_2021_CVPR}  & 0.19 & 0.04  & 0.10 & 0.02 & 0.07 & 0.01 & 0.13 & 0.03  \\
        F-Clip(HR)\cite{DAI20221} & \textcolor{red}{\textbf{0.60}} & \textcolor{red}{\textbf{0.42}}  & 0.25 & 0.17 & 0.21 & \textcolor{green}{\textbf{0.14}} & 0.35 & \textcolor{blue}{\textbf{0.25}}  \\
        HAWPv3\cite{xue2020holistically}  & 0.22 & 0.07  & 0.15 & 0.05 & 0.10 & 0.03 & 0.18 & 0.05  \\
        \bottomrule
    \end{tabular}
    }
    \end{subtable}
    \hfill
    \begin{subtable}{0.49\textwidth}
    \centering
    \caption{Metrics on the YULD\cite{cho2017novel}.}
    \label{tabel:2}
    \resizebox{\textwidth}{!}{
    \centering
    \begin{tabular}{ccccccccc}
        \toprule
        \multirow{2}{*}{Method} & \multicolumn{2}{c}{$\rm{AP\uparrow}$} & \multicolumn{2}{c}{$\rm{AR\uparrow}$} & \multicolumn{2}{c}{$\rm{IoU\uparrow}$} & \multicolumn{2}{c}{$\rm{F-Score\uparrow}$} \\ 
        \cline{2-9} 
        & $\rm Con_1$ & $\rm Con_2$ & $\rm Con_1$ & $\rm Con_2$ & $\rm Con_1$ & $\rm Con_2$ & $\rm Con_1$ & $\rm Con_2$\\ 
        \midrule
        LSD\cite{von2008lsd}  & 0.29 & 0.16 & 0.32 & 0.17 & 0.18 & 0.09 & 0.30 & 0.16 \\
        EDLines\cite{akinlar2011edlines}  & 0.26 & 0.12 & 0.26 & 0.11 & 0.15 & 0.06 & 0.26 & 0.12 \\
        Linelet\cite{cho2017novel}  & 0.30 & 0.15 & \textcolor{green}{\textbf{0.40}} & \textcolor{red}{\textbf{0.20}} & \textcolor{blue}{\textbf{0.21}}& \textcolor{green}{\textbf{0.10}} & \textcolor{blue}{\textbf{0.35}} & \textcolor{green}{\textbf{0.17}} \\
        MCMLSD\cite{almazan2017mcmlsd} & 0.23 & 0.10 & 0.33 & 0.14 & 0.16 & 0.06 & 0.27 & 0.11 \\
        AG3Line\cite{ZHANG2021107834}  & 0.34 & 0.14 & \textcolor{red}{\textbf{0.43}} & \textcolor{blue}{\textbf{0.18}} & \textcolor{red}{\textbf{0.23} }& 0.09 & \textcolor{red}{\textbf{0.38}} & 0.16 \\
        ELSED\cite{suarez2022elsed}  & \textcolor{green}{\textbf{0.36}} & \textcolor{blue}{\textbf{0.16}} & \textcolor{blue}{\textbf{0.36}} & 0.16 & \textcolor{green}{\textbf{0.22}} & 0.09 & \textcolor{green}{\textbf{0.36}} & 0.16\\
        \textbf{AAGLSD}(Ours)  & \textcolor{blue}{\textbf{0.36}} & \textcolor{green}{\textbf{0.21}} & 0.32 & \textcolor{green}{\textbf{0.19}} & 0.20 & \textcolor{red}{\textbf{0.11}} & 0.34 & \textcolor{red}{\textbf{0.20}} \\
        \hline
        \hline
        L-CNN\cite{9008267} & 0.35 & 0.18  & 0.27 & 0.14 & 0.18 & 0.08 & 0.31 & 0.16 \\
        SOLD$\rm{^2}$\cite{Pautrat_Lin_2021_CVPR} & 0.26 & 0.06  & 0.14 & 0.03 & 0.10 & 0.02 & 0.18 & 0.04  \\
        F-Clip(HR)\cite{DAI20221} & \textcolor{red}{\textbf{0.39}} & \textcolor{red}{\textbf{0.27}} & 0.17 & 0.12 & 0.13 & \textcolor{blue}{\textbf{0.09}} & 0.24 & \textcolor{blue}{\textbf{0.17}}\\
        HAWPv3\cite{xue2020holistically} & 0.27 & 0.08  & 0.20 & 0.06 & 0.13 & 0.03 & 0.23 & 0.07 \\
        \bottomrule
    \end{tabular}
}
    \end{subtable}
\end{table}

\subsection{Hyper-parameters Setting}
\label{ssec:parameter setting}
Following \cite{desolneux2002dequantizing,von2008lsd}, we set $T_{aligned}=\frac{\pi}{8}$, and $T_{mag}=5.22$. The magnitude difference threshold $T_{anchor}=3.0$ and the distance tolerance $T_{dist}=1.5$. In the process of validation and merging, the length threshold $T_{len}=9$, the minimum number of AAGs $N_{aag}=3$, the anchor density threshold $\rho_1=0.5$, the alignment density threshold $\rho_2=0.5$, the perpendicular center distance threshold between line segments $D_{c}=1.5$, the angular difference threshold $D_a=\frac{\pi}{36}$, and the endpoint distance threshold $D_{e}=9$.

\begin{table}[tb]
\caption{Repeatability of line segment detectors in different illuminations.}
\label{table:rep}
\centering
    \begin{tabular}{ccccccc}
            \toprule
            \multirow{3}{*}{Method} & \multicolumn{6}{c}{$\rm{repeatability\uparrow}$}\\ 
            \cline{2-7}
            & \multicolumn{3}{c}{$\lambda_{area}=0.75,D_a=\frac{\pi}{60}$} & \multicolumn{3}{c}{$\lambda_{area}=0.9,D_a=\frac{\pi}{60}$}\\
            & \multicolumn{1}{c}{$D_c=3.0$,} & \multicolumn{1}{c}{$D_c=2.0$,} & \multicolumn{1}{c}{$D_c=1.0$ } & \multicolumn{1}{c}{$D_c=3.0$,} & \multicolumn{1}{c}{$D_c=2.0$,} & \multicolumn{1}{c}{$D_c=1.0$ } \\ 
            \midrule
            LSD\cite{von2008lsd}  & 0.309 & 0.291 & 0.259 & 0.181 & 0.172 & \textbf{0.158}\\
            EDLines\cite{akinlar2011edlines}  & 0.266 & 0.245 & 0.216 & 0.142 & 0.132 & 0.121\\
            AAGLSD(Ours)  & \textbf{0.324} & \textbf{0.300} & \textbf{0.266} & \textbf{0.186} & \textbf{0.173} & 0.157\\
            \bottomrule
    \end{tabular}
\end{table}

    \subsection{Experiments Analysis}
\label{ssec:experiments analysis}
\subsubsection{Evaluation on YorkUrbanDB}

In addition to handcrafted methods, we evaluate supervised learning methods (L-CNN\cite{9008267} and F-Clip\cite{DAI20221}) and self-supervised learning methods (SOLD$\rm{^2}$\cite{Pautrat_Lin_2021_CVPR} and HAWPv3\cite{8575521}). All these methods are then inferred on both the YUD\cite{coughlan2003manhattan} and the YULD\cite{cho2017novel}. We use LSD\cite{von2008lsd} and EDLines\cite{akinlar2011edlines} in OpenCV\cite{opencv_library}, while the other methods adopt official implementations. We visualize qualitative results in the supplementary material.

To ensure that only predictions highly consistent with ground truths are considered true positives (i.e. midpoint offset and angle difference are very small), $D_c$ and $D_a$ are strictly set to \textbf{1.0} and $\bm{\frac{\pi}{60}}$. \Cref{fig:curves} presents the performance curves for different line segment detectors on the YUD\cite{coughlan2003manhattan} and the YULD\cite{cho2017novel} in various $\lambda_{area}$. The results of quantitative experiments for $\lambda_{area}$ of \textbf{0.5} ($\rm Con_1$) and \textbf{0.9} ($\rm Con_2$) are detailed in \Cref{tabel:1} and \Cref{tabel:2}. 

In our experiments, F-Clip\cite{DAI20221} obtains the highest AP but a relatively low AR, since it filters out line segments with low confidence, retaining only those that are long. L-CNN\cite{9008267} achieves a good balance between precision and recall. However, a significant performance drop is observed on the YULD\cite{cho2017novel} for both approaches\cite{9008267,DAI20221}. It originates from their fully supervised training paradigm for wireframe parsing. Compared to the baseline method, LSD\cite{von2008lsd}, the other two self-supervised methods\cite{xue2020holistically,Pautrat_Lin_2021_CVPR} exhibit poor performance due to their pseudo-label generator trained on a synthetic dataset consisting of several primitives (e.g., chessboard), resulting in incomplete line segments. The results suggest that handcrafted line segment detectors with inductive bias exhibit better generalization ability in structured scenes, and self-supervised learning-based methods need more in-depth research.

Among handcrafted detectors, our AAGLSD has one of the best performances in both precision and recall, especially at high $\lambda_{area}$ (see \Cref{fig:curves}). It is also observed that ELSED\cite{suarez2022elsed} exhibits a performance comparable to our AAGLSD and can even surpass it under less stringent conditions. Since ELSED\cite{suarez2022elsed} mainly refines the edge drawing process without incorporating extra cues related to line segment information, it struggles to accurately and completely detect line segments under more challenging conditions (e.g. $\rm Con_2$). The proposed AAGLSD maintains a consistently high AP in various $\lambda_{area}$ with an advanced AR. As $\lambda_{area}$ increases, it demonstrates enhanced competitiveness, such as $\rm Con_2$. This is attributed to AAGLSD's ability to assign distinct saliency levels to different pixels, which facilitates the identification of line segments that are more likely to be integral parts of line structures. During the anchor-linking process, AAGLSD employs groups of pixels as cues, diverging from conventional approaches that rely solely on individual pixels. This strategy not only enhances the accuracy of linking orientation, but also avoids link termination caused by pixel shifts in rasterized line segments. Consequently, AAGLSD is capable of detecting line segments more completely than other handcrafted methods. It achieves an excellent trade-off between precision and recall, attaining the SOTA F-Score when $\lambda_{area}$ exceeds \textbf{0.8} and \textbf{0.825} on the YUD\cite{coughlan2003manhattan} and YULD\cite{cho2017novel} in our experiments, respectively.

\subsubsection{Evaluation on HPatches}
We report \textit{repeatability} of AAGLSD with the two most popular handcrafted methods, LSD\cite{von2008lsd} and EDLines\cite{akinlar2011edlines}, in the illumination subset of HPatches\cite{hpatches_2017_cvpr} to evaluate robustness, as shown in \Cref{table:rep}. It demonstrates that our method could robustly detect line segments in various illuminations. We visualize the line segments detected by our method in the supplementary material.

\section{Conclusion}
\label{sec:conclusion}
This paper introduces a line segment detector, AAGLSD, which employs a hierarchical anchor extraction strategy and frames line segment detection as an anchor-linking process. By selecting pixels with a high probability of being part of line segments, it yields more complete predictions. Experimental validation demonstrates that AAGLSD achieves an excellent trade-off between precision and recall metrics under challenging conditions. Furthermore, it attains robust detection performance against illumination variations.

%
%
%
\bibliographystyle{splncs04}
\bibliography{refs}

\end{document}